\documentclass[letterpaper]{article}
\usepackage{aaai2027} 

\usepackage[hyphens]{url}
\usepackage{graphicx}
\usepackage{natbib}
\usepackage{caption}
\usepackage{algorithm}
\usepackage{algorithmic}
\usepackage{amsmath}
\usepackage{amssymb}

\usepackage{newfloat}
\usepackage{listings}
\DeclareCaptionStyle{ruled}{
    labelfont=normalfont,
    labelsep=colon,
    strut=off
}
\floatstyle{ruled}
\newfloat{listing}{tb}{lst}{}
\floatname{listing}{Listing}

\usepackage{booktabs}
\usepackage{multirow}
\usepackage{capt-of}

\title{
    Hyper-RED: Scalable Event Pre-training via Semantic Hypergraph Distillation
}

\author{
    Meisen Wang\textsuperscript{\rm 1},
    Zhiqiang Tian\textsuperscript{\rm 1,\rm 2},
    Wei Bao\textsuperscript{\rm 3,\rm 4},
    Chengjie Wang\textsuperscript{\rm 5},\\
    Shaoyi Du\textsuperscript{\rm 2},
    Siqi Li\textsuperscript{\rm 3,\rm 4},
}

\affiliations{
    \textsuperscript{\rm 1}School of Software Engineering,
    Xi'an Jiaotong University, Xi'an 710049, China\\
    \textsuperscript{\rm 2}National Key Laboratory of
    Human-Machine Hybrid Augmented Intelligence,\\
    National Engineering Research Center for
    Visual Information and Applications,\\
    and Institute of Artificial Intelligence and Robotics,\\
    Xi'an Jiaotong University, Xi'an 710049, China\\
    \textsuperscript{\rm 3}BNRist, THUIBCS, BLBCI,
    School of Software, Tsinghua University, Beijing 100084, China\\
    \textsuperscript{\rm 4}Yangtze Delta Region Institute,
    Tsinghua University, Jiaxing 314006, China\\
    \textsuperscript{\rm 5}College of Grassland Science,
    Inner Mongolia Agricultural University, Hohhot 010018, China\\
    3125158005@stu.xjtu.edu.cn,
    denghao293@stu.xjtu.edu.cn\\
    zhiqiangtian@xjtu.edu.cn,
    dushaoyi@xjtu.edu.cn,
    nmgcjwang3@imau.edu.cn\\
    \{baoweivvv,lisiqi19971013,kevin.gaoy\}@gmail.com
}

\begin{document}

\maketitle

\begin{abstract}

Event cameras have shown great potential for robust visual perception, yet scaling event representation learning remains challenging due to the scarcity of large-scale annotated event data.
Pretrained image models provide scalable semantic supervision, but existing image-to-event methods rely on rigid pixel-wise or token-wise alignment that overlooks modality discrepancies in texture, density, and appearance, potentially causing semantic collapse and limiting transferability.
To address this issue, we propose Hyper-RED, a simple, painless, and scalable image-to-event pretraining framework that transfers high-order semantic structures from images to events.
Hyper-RED uses hypergraphs to model and align high-order semantic associations among multiple image and event tokens, enabling cross-modal knowledge transfer while accommodating modality-specific differences rather than enforcing rigid one-to-one correspondence.
Specifically, given a paired event--image sample, Hyper-RED leverages DINOv3 to extract spatial token representations and constructs image, event, and cross-modal semantic hypergraphs, where each hyperedge connects multiple semantically correlated tokens.
We further introduce a hypergraph relational distillation loss that imposes complementary intra- and cross-modal constraints, enabling the event encoder to inherit image-derived semantic organization while preserving local relational consistency and event-specific characteristics.
Experiments on three tasks across five event datasets demonstrate consistent scaling from ViT-S to ViT-L and state-of-the-art performance (Fig.~\ref{fig:radar}).

\begin{figure}[t]
    \centering\includegraphics[width = 0.9\linewidth]{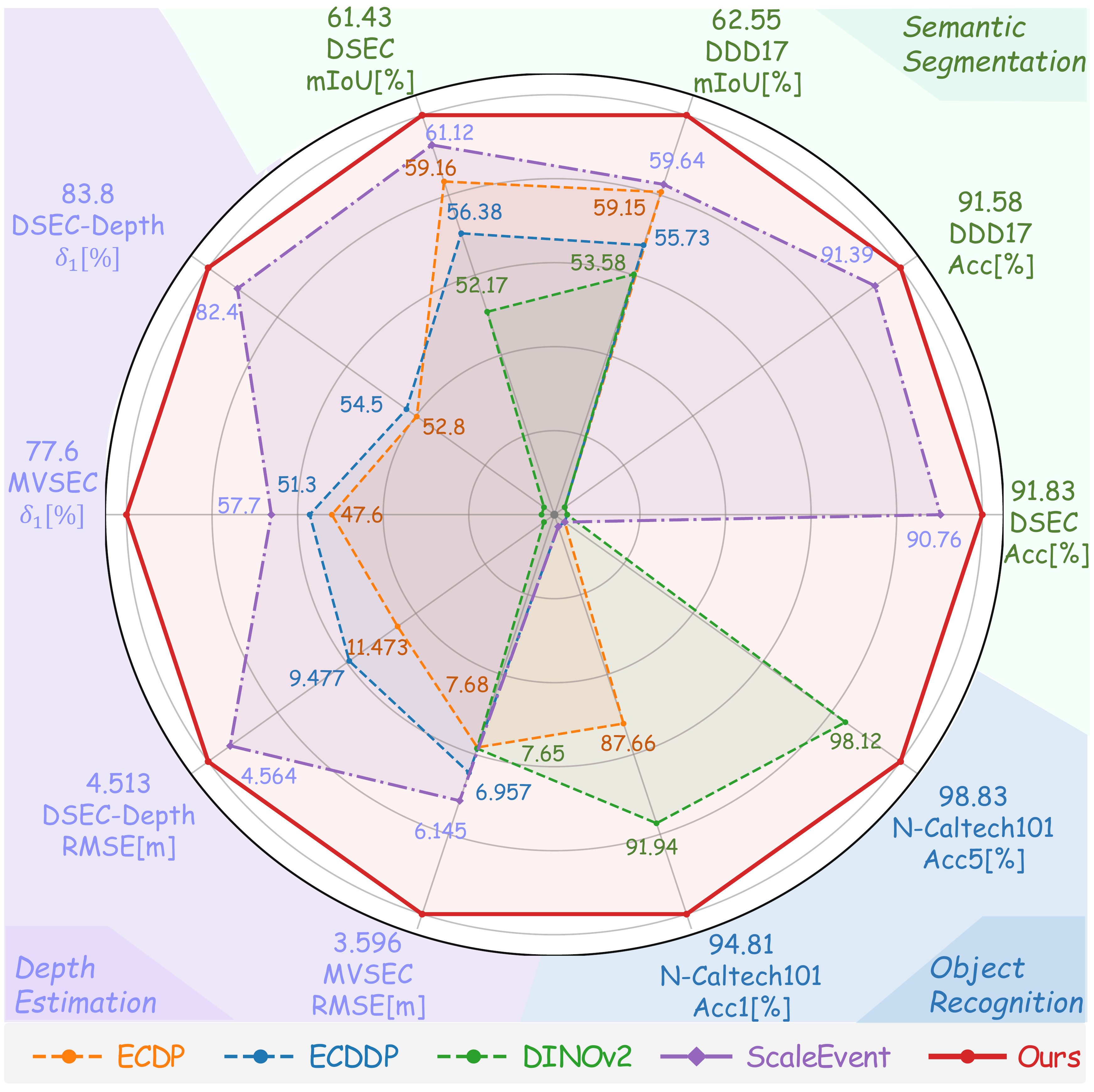}
    \caption{Overall comparison across ten metrics on various datasets. With the same backbone model, our method demonstrates superior and consistent performance across all tasks.}
    \label{fig:radar}
\end{figure}

\end{abstract}

\section{Introduction}

Event cameras asynchronously record per-pixel brightness changes with high temporal resolution, high dynamic range, low latency, and low power consumption, making them well suited to fast motion and extreme illumination~\cite{gallego2020event}.
These advantages have motivated event-based representation learning for a wide range of visual perception tasks, including recognition~\cite{yang2025ezsr}, semantic segmentation~\cite{kong2024openess,jing2024hpl}, depth estimation~\cite{zhu2023self,bartolomei2025depth}, and object detection~\cite{peng2024scene,chen2025event}. However, learning transferable event representations at scale remains challenging because event streams are sparse, asynchronous, and lack rich texture cues, while large-scale annotated event datasets remain far less abundant than image datasets. In contrast to event-based vision, the image domain has benefited from powerful pretrained models and visual foundation models~\cite{caron2021emerging, oquab2023dinov2, simeoni2025dinov3, radford2021learning, he2022masked, guo2026phaselora}, which provide strong semantic representations across diverse downstream tasks. Transferring these semantic priors to event encoders therefore provides a scalable alternative to annotation-intensive event pre-training.

\begin{figure}[t]
    \centering\includegraphics[width = \linewidth]{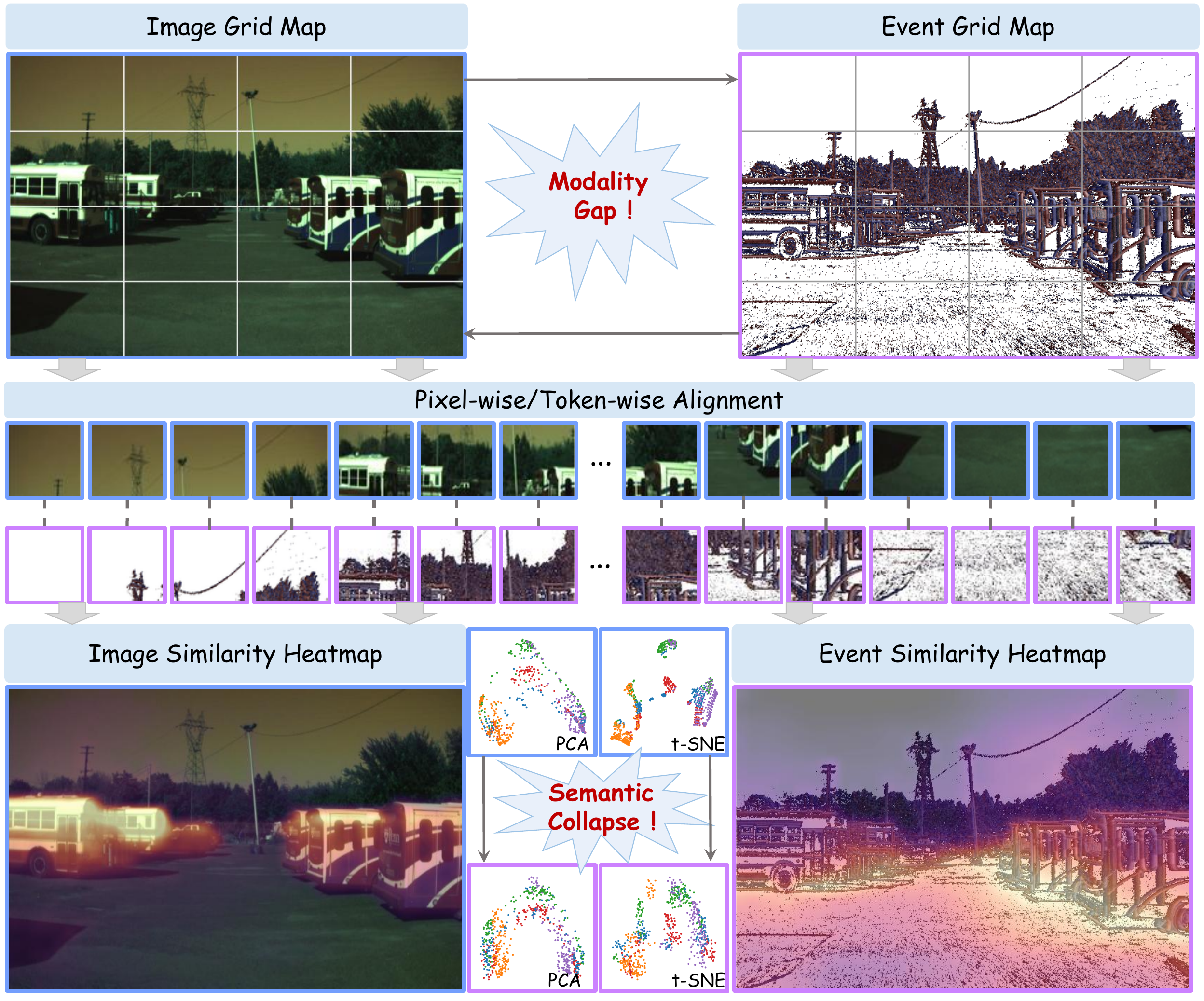}
    \caption{Representation analysis under rigid image--event alignment. We train an event encoder through rigid alignment with a frozen image encoder, and compare their similarity heatmaps on paired inputs. PCA and t-SNE further visualize the feature distributions of the two encoders.}
    \label{fig:semantic_collapse}
\end{figure}

Recent event pretraining methods have alleviated the scarcity of event annotations by leveraging pretrained image models for event representation learning~\cite{yang2023event,yang2024event, liang2025efficient, wu2025cm3ae, cao2026generative}. Although these methods have achieved promising performance, existing distillation paradigms primarily rely on embedding-level, pixel-wise, or token-wise correspondence between paired event and image inputs. Such a design implicitly assumes that spatially corresponding features should encode consistent semantics across modalities. However, this assumption becomes fragile under the inherent image--event modality gap, where image representations are strongly shaped by dense appearance cues, while event representations are derived from sparse motion- and contrast-sensitive responses. As shown in Fig.~\ref{fig:semantic_collapse}, compared with the compact similarity patterns of the image encoder, those of the rigidly aligned event encoder become noticeably more diffuse, accompanied by reduced class separability. This observation reveals that rigid alignment fails to preserve semantic organization across the modality gap, posing a key bottleneck to scalable image-to-event pre-training. ScaleEvent~\cite{chen2026scaling} recognizes this limitation and alleviates rigid point-wise matching through graph-based structural alignment. However, its pairwise formulation decomposes the multi-token associations underlying high-order semantics into isolated relations, potentially causing semantic information loss during cross-modal transfer. Motivated by this observation, we rethink image-to-event distillation from the perspective of high-order semantic relation alignment, aiming to transfer modality-invariant semantic structures beyond isolated spatial correspondence and pairwise graph relations.

To address this limitation, we propose Hyper-RED, a simple, painless and scalable image-to-event pretraining distillation paradigm that transfers high-order semantic structures across modalities. By exploiting the multi-node relational capacity of hypergraphs, Hyper-RED captures shared semantics and aligns high-order semantic associations among multiple image and event tokens, thereby narrowing the cross-modal representation gap without enforcing rigid one-to-one correspondence. Given paired event--image samples, Hyper-RED extracts spatial token representations using a frozen DINOv3 image teacher and a trainable event encoder, and constructs image, event, and cross-modal semantic hypergraphs. Each hyperedge groups semantically correlated tokens to represent region-level dependencies beyond isolated point-wise correspondence. Based on these hypergraphs, we introduce a hypergraph relational distillation loss with complementary intra-modal and cross-modal constraints, allowing the event encoder to inherit image-derived semantic organization while preserving event-specific characteristics.  Experiments on three downstream tasks across five datasets show that Hyper-RED scales consistently from ViT-S to ViT-L and achieves state-of-the-art overall performance in object recognition, semantic segmentation, and depth estimation. 

Our contributions are summarized as follows:

\begin{itemize}

\item We propose Hyper-RED, a scalable image-to-event pre-training framework that distills high-order semantic structures from a frozen visual foundation model through semantic hypergraphs, moving beyond point-wise matching and pairwise relations.

\item We revisit semantic collapse arising from image--event mismatch under rigid alignment and introduce a hypergraph relational distillation loss that aligns hyperedge memberships and prototypes through intra- and cross-modal constraints, thereby enabling reliable semantic transfer while preserving event-specific characteristics.

\item We demonstrate state-of-the-art overall performance on five benchmarks across three tasks, with consistent scaling from ViT-S to ViT-L and strong transferability under linear probing, few-shot fine-tuning, and full supervision.

\end{itemize}

\section{Related Work}

\subsection{Event-Based Representation Pre-training}

\noindent\textbf{Event-Only Self-Supervision.}
The sparsity and asynchronous nature of event streams, together with the scarcity of large-scale annotations, have motivated self-supervised pre-training directly on unlabeled event data. Masked Event Modeling~\cite{klenk2024masked} learns transferable representations by reconstructing masked event inputs, whereas TESPEC~\cite{mohammadi2025tespec} extends this paradigm to long event sequences and recurrent encoders. By deriving supervisory signals entirely from event streams, these methods eliminate the need for paired event--image data or manual annotations during pre-training.

\noindent\textbf{Cross-Modal Knowledge Transfer.}
Another line of research transfers semantic priors from large-scale pretrained image models to event encoders. An early representative study distilled multi-level image features into event networks for recognition and optical flow~\cite{deng2021learning}. ECDP~\cite{yang2023event} and ECDDP~\cite{yang2024event} subsequently transferred image knowledge to event encoders for generic representation learning and dense prediction, respectively. Subsequent studies extended this paradigm through prompt fusion~\cite{liang2025efficient}, cross-modal masked modeling~\cite{wu2025cm3ae}, and generative pre-training~\cite{cao2026generative}. The Cross-Modal Event Encoder~\cite{jeong2026cross} further transfers image--text knowledge to event streams while retaining text-aligned zero-shot capabilities. ScaleEvent~\cite{chen2026scaling} advances beyond direct feature matching by introducing graph-based alignment of pairwise token relations.

\noindent\textbf{Remark.}
Event-only self-supervision eliminates the need for paired event--image data, but its effectiveness depends heavily on the scale and diversity of event data and on carefully designed pretext tasks. Cross-modal knowledge transfer instead allows event encoders to inherit rich semantic priors from visual foundation models, reducing the burden of discovering high-level semantics solely from sparse event streams. Nevertheless, existing image-to-event distillation methods primarily rely on point-wise feature imitation, task-specific proxy supervision, or pairwise relational constraints. Such objectives are sensitive to the image--event modality gap and provide insufficient supervision for preserving region-level semantic organization. Scalable image-to-event pre-training therefore requires robust, task-agnostic alignment beyond isolated point-wise and pairwise relations.

\subsection{Hypergraph Representation Learning}

Hypergraphs generalize ordinary graphs by allowing a single hyperedge to connect multiple nodes, thereby representing group-wise dependencies beyond pairwise edges~\cite{gao2022hgnn+, gao2020hypergraph}. HGNN~\cite{feng2019hypergraph} introduced hypergraph convolution to encode complex data correlations during representation learning. DHGNN~\cite{jiang2019dynamic} subsequently learned and dynamically updated hypergraph structures from evolving features, while UniGNN~\cite{huang2021unignn} formulated a unified message-passing framework for graph and hypergraph neural networks. Beyond architectural design, hypergraphs have also been incorporated into learning objectives. In visual metric learning, HIST~\cite{lim2022hypergraph} employs semantic tuplets to capture multilateral sample-to-class relations that cannot be represented by independent pairwise constraints. Collectively, these studies establish hypergraphs as an effective mechanism for modeling high-order relations among multiple entities.

\noindent\textbf{Remark.}
Lessons from existing studies demonstrate that hypergraphs are effective at modeling high-order dependencies among multiple entities. However, they have primarily been used as prediction backbones or task-specific learning objectives within a single modality. Consequently, their potential as structured knowledge carriers between heterogeneous visual modalities remains underexplored. This gap motivates us to investigate semantic hypergraphs as structured carriers for image-to-event relational distillation.

\section{Preliminary}

\begin{figure*}[t]
    \centering
    \includegraphics[width=\linewidth]{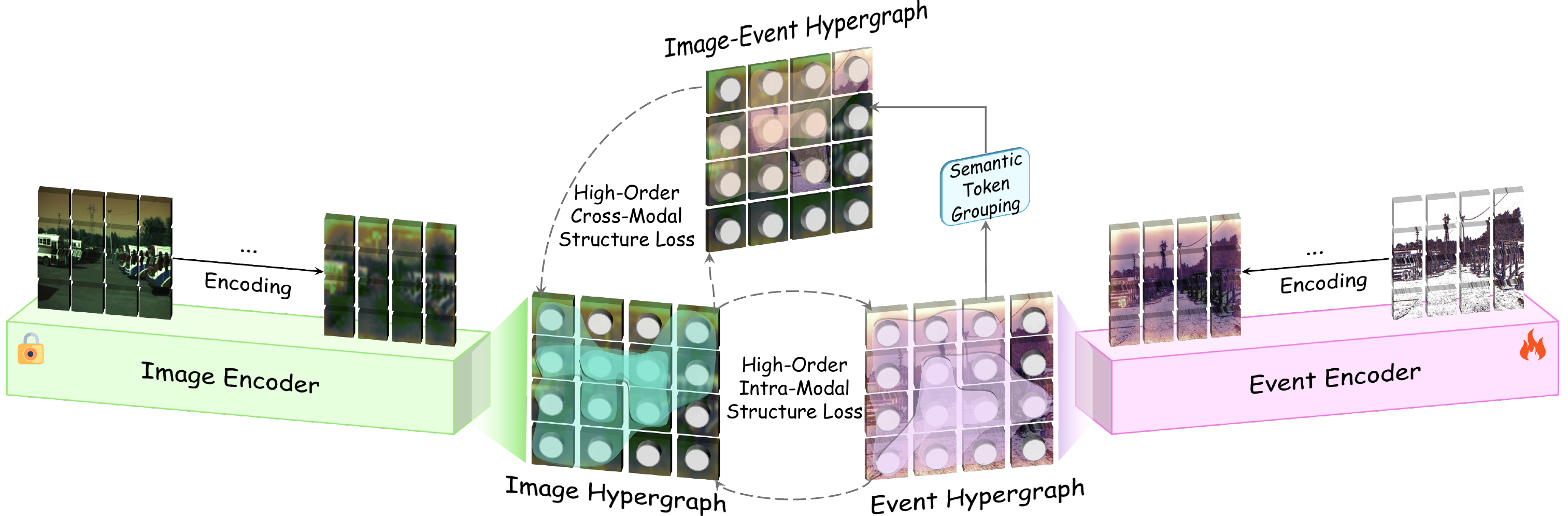}
    \caption{Overview of Hyper-RED. Hyper-RED mitigates semantic collapse from rigid one-to-one image--event alignment by constructing teacher-guided semantic hypergraphs and distilling high-order intra- and cross-modal relations, yielding transferable, modality-invariant event representations across encoder scales.}
    \label{fig:pipeline}
\end{figure*}

\noindent\textbf{Synchronized Event--Image Data.}
Let $\mathbf{E}$ denote an event representation constructed from an event stream and $\mathbf{I}$ its spatially and temporally aligned image. Although they describe the same scene, the two modalities exhibit substantially different sensing characteristics: images contain dense appearance and texture cues, whereas event data capture sparse brightness changes caused by motion or illumination variation. This modality gap makes isolated one-to-one token correspondence insufficient for transferring the semantic organization learned by an image foundation model.

\noindent\textbf{Cross-Modal Distillation.}
Given an aligned pair $(\mathbf{E},\mathbf{I})$, a trainable event encoder $F_{\theta_e}$ and a frozen image teacher $G_{\theta_i}$ extract spatial token features
\begin{equation}
    \mathbf{K}=F_{\theta_e}(\mathbf{E}),
    \qquad
    \mathbf{Q}=G_{\theta_i}(\mathbf{I}),
    \label{eq:cross_modal_features}
\end{equation}
where $\mathbf{K},\mathbf{Q}\in\mathbb{R}^{N\times D}$ contain $N$ spatial tokens of dimension $D$. The two encoders share the same spatial token layout, enabling the teacher to provide aligned supervision. Conventional distillation directly aligns spatially corresponding tokens in $\mathbf{K}$ and $\mathbf{Q}$. Beyond this one-to-one alignment, Hyper-RED transfers the semantic organization among multiple tokens, allowing the event encoder to inherit region-level structure from the teacher.

\noindent\textbf{Hypergraph Representation.}
A hypergraph $\mathcal{G}=(\mathcal{V},\mathcal{E}_{h},\mathbf{H})$ comprises a node set $\mathcal{V}$, a hyperedge set $\mathcal{E}_{h}$, and an incidence matrix $\mathbf{H}$. Unlike an ordinary graph edge that relates two nodes, a hyperedge jointly connects multiple nodes and captures group-wise dependencies. In our formulation, each node corresponds to a spatial token, each hyperedge represents a semantic group, and the incidence weights quantify the relative contributions of its connected tokens.

\section{Methodology}
\label{sec:method}

Our goal is to learn fine-grained event representations by transferring high-order semantic structures from a frozen pretrained DINOv3~\cite{simeoni2025dinov3} image teacher using synchronized event--image data without manual annotations. To this end, we propose Hyper-RED, a teacher-guided semantic hypergraph distillation framework. As illustrated in Fig.~\ref{fig:pipeline}, Hyper-RED organizes semantically correlated tokens into teacher-defined hyperedges and transfers their group-wise organization to the event encoder. It introduces two complementary objectives: a high-order intra-modal structure loss that aligns image and event token organization, and a high-order cross-modal structure loss that establishes correspondence between image semantic anchors and event-token groups. Because the construction operates on normalized token affinities, the formulation is agnostic to feature dimension and applies uniformly to ViT-S/B/L encoders without scale-specific modifications.

\subsection{Semantic Hypergraph Construction}
\label{sec:hypergraph_construction}

\noindent\textbf{Teacher-Guided Semantic Grouping.}
We first $\ell_2$-normalize each token in $\mathbf{Q}$ and $\mathbf{K}$ along the feature dimension and reuse the same notation for the normalized matrices. We then construct image--image, event--event, and image--event affinity matrices as
\begin{equation}
    \mathbf{S}^{\mathrm{I}}=[\mathbf{Q}\mathbf{Q}^{\top}]_{+},
    \quad
    \mathbf{S}^{\mathrm{E}}=[\mathbf{K}\mathbf{K}^{\top}]_{+},
    \quad
    \mathbf{S}^{\mathrm{X}}=[\mathbf{Q}\mathbf{K}^{\top}]_{+},
    \label{eq:semantic_affinities}
\end{equation}
where $[\cdot]_{+}$ clips negative similarities to zero. The teacher-derived matrix $\mathbf{S}^{\mathrm{I}}$ provides a stable structural reference, while $\mathbf{S}^{\mathrm{E}}$ and $\mathbf{S}^{\mathrm{X}}$ characterize the corresponding event--event and image--event relations, respectively. In particular, the $a_e$-th row of $\mathbf{S}^{\mathrm{X}}$ measures the associations between image anchor $\mathbf{q}_{a_e}$ and all event tokens, allowing the teacher anchor to directly query the event representation.

We rank image tokens by their total affinities to all other tokens and select the $M$ highest-scoring ones as semantic anchors, whose indices form
$\mathcal{A}=\{a_e\}_{e=1}^{M}$. For each anchor $a_e$, its $k$ most correlated image token indices form the member set
\begin{equation}
    \mathcal{V}_e
        =\operatorname{TopKIndices}_{k}
        \left(\left\{S^{\mathrm{I}}_{a_ej}\right\}_{j=1}^{N}\right).
    \label{eq:semantic_members}
\end{equation}
Each hyperedge thus groups an anchor with multiple tokens from a coherent teacher-defined semantic region. Because the two encoders share the same spatial token layout, the teacher-selected anchor and member indices define a common topology across all branches. This design ensures that the losses compare branch-specific relation strengths within identical semantic groups rather than mismatched neighborhoods.

\noindent\textbf{Soft Hyperedge Representation.}
Although the three branches share a topology, their member contributions depend on branch-specific affinities. For branch $r\in\{\mathrm{I},\mathrm{E},\mathrm{X}\}$, the soft membership of token $i$ in hyperedge $e$ is
\begin{equation}
    H^{r}_{ie}
    =\frac{\exp(S^{r}_{a_ei}/\tau)}
    {\sum_{j\in\mathcal{V}_e}\exp(S^{r}_{a_ej}/\tau)},
    \qquad i\in\mathcal{V}_e ,
    \label{eq:soft_membership}
\end{equation}
where $\tau=0.07$ controls the distribution concentration, $H^{r}_{ie}=0$ for $i\notin\mathcal{V}_e$, and $\mathbf{H}^{r}\in\mathbb{R}^{N\times M}$. The matrices $\mathbf{H}^{\mathrm{I}}$, $\mathbf{H}^{\mathrm{E}}$, and $\mathbf{H}^{\mathrm{X}}$ encode teacher grouping, event-token organization, and cross-modal association, respectively.

Membership distributions characterize the internal organization of a hyperedge but do not explicitly summarize its semantic content. We therefore aggregate the member tokens into a normalized prototype:
\begin{equation}
    \mathbf{p}^{r}_e
    =\mathrm{Norm}\left(
    \sum_{i\in\mathcal{V}_e}H^{r}_{ie}\mathbf{u}^{r}_i
    \right),
    \label{eq:semantic_prototype}
\end{equation}
where $\mathbf{u}^{\mathrm{I}}_i=\mathbf{q}_i$ and $\mathbf{u}^{\mathrm{E}}_i=\mathbf{u}^{\mathrm{X}}_i=\mathbf{k}_i$. Here, $\mathbf{q}_i$ and $\mathbf{k}_i$ are normalized image and event tokens, respectively, and $\mathrm{Norm}$ denotes $\ell_2$ normalization. The membership distribution and prototype therefore encode the internal composition and aggregated semantics of each hyperedge, respectively.

\begin{table*}[t]
    \centering
    \begingroup
    \small
    \setlength{\tabcolsep}{0.25mm}
    \begin{tabular*}{\textwidth}{
        @{\extracolsep{\fill}}
        lllcccccccccccc
        @{}
    }
        \toprule
        \multirow{2}{*}{Method}
        & \multirow{2}{*}{Venue}
        & \multirow{2}{*}{Backbone}
        & \multicolumn{6}{c}{MVSEC-Depth}
        & \multicolumn{6}{c}{DSEC-Depth} \\
        \cmidrule(lr){4-9}
        \cmidrule(lr){10-15}
        & & &
        $\delta_1\uparrow$ &
        $\delta_2\uparrow$ &
        $\delta_3\uparrow$ &
        AbsRel$\downarrow$ &
        RMSE$\downarrow$ &
        RMSE$_{\log}\downarrow$ &
        $\delta_1\uparrow$ &
        $\delta_2\uparrow$ &
        $\delta_3\uparrow$ &
        AbsRel$\downarrow$ &
        RMSE$\downarrow$ &
        RMSE$_{\log}\downarrow$ \\
        \midrule

        \multicolumn{15}{l}{
            \textit{RGB Initialization$^{\dagger}$}
        } \\

        E2Depth
        & 3DV'20
        & ResNet-18
        & 0.432
        & 0.717
        & 0.868
        & 0.420
        & 7.268
        & 0.455
        & 0.409
        & 0.719
        & 0.891
        & 0.395
        & 13.258
        & 0.412 \\

        EReFormer
        & TCSVT'24
        & Swin-T
        & 0.391
        & 0.652
        & 0.810
        & 0.551
        & 8.373
        & 0.523
        & 0.524
        & 0.824
        & 0.945
        & 0.297
        & 11.608
        & 0.334 \\

        \midrule
        \multicolumn{15}{l}{
            \textit{Event Pretraining$^{\dagger}$}
        } \\

        ECDP
        & ICCV'23
        & ViT-S/16
        & 0.476
        & 0.772
        & 0.863
        & 0.496
        & 7.680
        & 0.506
        & 0.528
        & 0.818
        & 0.938
        & 0.324
        & 11.473
        & 0.376 \\

        ECDDP
        & ECCV'24
        & ViT-S/16
        & 0.513
        & 0.762
        & 0.871
        & 0.428
        & 6.957
        & 0.469
        & 0.545
        & 0.857
        & 0.959
        & 0.263
        & 9.477
        & 0.294 \\

        DepthAnyEvent-R
        & ICCV'25
        & ViT-S/16
        & 0.489
        & 0.751
        & 0.878
        & 0.365
        & 6.465
        & 0.483
        & 0.691
        & 0.930
        & 0.981
        & 0.191
        & 8.880
        & 0.266 \\

        ScaleEvent
        & CVPR'26
        & ViT-L/16
        & 0.625
        & 0.834
        & 0.934
        & 0.268
        & 5.554
        & 0.343
        & \underline{0.896}
        & \underline{0.983}
        & \underline{0.997}
        & \underline{0.101}
        & \underline{3.694}
        & \underline{0.144} \\

        \midrule

        Ours
        & --
        & ViT-S/16
        & 0.776
        & 0.940
        & 0.982
        & 0.165
        & 3.596
        & 0.214
        & 0.838
        & 0.965
        & 0.992
        & 0.123
        & 4.513
        & 0.167 \\

        Ours
        & --
        & ViT-B/16
        & \underline{0.804}
        & \underline{0.943}
        & \underline{0.982}
        & \underline{0.155}
        & \underline{3.449}
        & \underline{0.205}
        & 0.864
        & 0.974
        & 0.995
        & 0.111
        & 4.193
        & 0.158 \\

        \textbf{Ours}
        & --
        & ViT-L/16
        & \textbf{0.807}
        & \textbf{0.947}
        & \textbf{0.983}
        & \textbf{0.152}
        & \textbf{3.383}
        & \textbf{0.202}
        & \textbf{0.904}
        & \textbf{0.985}
        & \textbf{0.998}
        & \textbf{0.096}
        & \textbf{3.624}
        & \textbf{0.137} \\

        \bottomrule
    \end{tabular*}
    \endgroup
    \caption{Comparison of monocular depth estimation performance on the MVSEC-Depth and DSEC-Depth. The best and second-best results are shown in \textbf{bold} and \underline{underline}, respectively. $^{\dagger}$ denotes fully supervised training.}   
    \label{tab:event_depth_results}
\end{table*}

\subsection{High-Order Structure Distillation}
\label{sec:high_order_distillation}

The image teacher defines the topology and target organization of the semantic hypergraph. We transfer this knowledge by aligning the teacher hypergraph with the event--event and image--event hypergraphs. All divergences are averaged over the $M$ constructed hyperedges, and teacher-derived memberships and prototypes are detached during optimization.

\noindent\textbf{High-Order Intra-Modal Structure Loss.}
The event--event branch models the internal semantic geometry of event tokens. We align its membership distribution $\mathbf{H}^{\mathrm{E}}$ with the teacher distribution $\mathbf{H}^{\mathrm{I}}$ using KL divergence and align their hyperedge prototypes using cosine distance:
\begin{equation}
    \mathcal{L}_{\mathrm{HOI}}
        =\mathrm{KL}(\mathbf{H}^{\mathrm{I}}\|\mathbf{H}^{\mathrm{E}})
        +\left[1-\cos(\mathbf{P}^{\mathrm{I}},\mathbf{P}^{\mathrm{E}})\right],
    \label{eq:hoi_loss}
\end{equation}
where $\mathbf{P}^{r}=[\mathbf{p}^{r}_1,\ldots,\mathbf{p}^{r}_M]^{\top}\in\mathbb{R}^{M\times D}$ stacks the $M$ prototypes. Here and below, the cosine term averages similarities between corresponding prototypes over all hyperedges. The membership term transfers the relative roles of tokens within each semantic group, while the prototype term preserves its group-level representation. It is termed intra-modal because the two hypergraphs are independently constructed from image--image and event--event relations before structural alignment.

\noindent\textbf{High-Order Cross-Modal Structure Loss.}
Internal event-token consistency alone does not ensure alignment with the image semantic space. We therefore use the image--event branch derived from cross-modal affinities and impose
\begin{equation}
    \mathcal{L}_{\mathrm{HOC}}
        =\mathrm{KL}(\mathbf{H}^{\mathrm{I}}\|\mathbf{H}^{\mathrm{X}})
        +\left[1-\cos(\mathbf{P}^{\mathrm{I}},\mathbf{P}^{\mathrm{X}})\right].
    \label{eq:hoc_loss}
\end{equation}
This loss transfers the teacher-defined semantic association between each image anchor and its event-token group. Consequently, $\mathcal{L}_{\mathrm{HOI}}$ preserves the organization among event tokens, whereas $\mathcal{L}_{\mathrm{HOC}}$ anchors that organization to the teacher. Their combination provides complementary high-order supervision beyond isolated token-wise correspondence and pairwise relations.

\noindent\textbf{Overall Objective.}
The complete objective is
\begin{equation}
    \mathcal{L}
    =
    \mathcal{L}_{\ell_1}
    +
    \mathcal{L}_{\mathrm{PRD}}
    +
    \lambda_{\mathrm{H}}
    \left(
    \mathcal{L}_{\mathrm{HOI}}
    +
    \mathcal{L}_{\mathrm{HOC}}
    \right),
    \label{eq:overall_objective}
\end{equation}
where $\lambda_{\mathrm{H}}=2$, and $\mathcal{L}_{\mathrm{PRD}}$ denotes the pairwise relational distillation loss \cite{chen2026scaling}.

\section{Experiments}

\subsection{Experimental Setup}

\noindent\textbf{Pre-training Datasets.}
To learn transferable event representations, we construct a large-scale pre-training corpus containing both real and simulated event data. The real-world subset comprises DSEC~\cite{gehrig2021dsec}, DDD17~\cite{li2019event}, MVSEC~\cite{zhu2018multivehicle}, CoeSot~\cite{tang2025revisiting}, VisEvent~\cite{wang2023visevent}, FEVD~\cite{kim2024frequency}, SEE-600K~\cite{lu2025see}, and HighREV~\cite{sun2023event}. The simulated subset is generated with v2e from Cityscapes~\cite{cordts2016cityscapes}, KITTI~\cite{geiger2013vision}, DAVIS 2017~\cite{pont20172017}, DECD~\cite{rebecq2019high}, and GoPro~\cite{nah2019ntire}. Following Ev-DTAD~\cite{wang2026rethinking}, we use HTA as the unified event representation and resize both modalities to $640 \times 480$. 

\begin{table}[t]
    \centering
    \begingroup
    \small
    \begin{tabular}{@{}lllcc@{}}
        \toprule
        \multirow{2}{*}{Method}
        & \multirow{2}{*}{Venue}
        & \multirow{2}{*}{Backbone}
        & \multicolumn{2}{c}{N-Caltech101} \\
        \cmidrule(lr){4-5}
        & & & acc1$\uparrow$ & acc5$\uparrow$ \\
        \midrule

        \multicolumn{5}{l}{\textit{Training from scratch}} \\
        ViT
        & ICLR'21
        & ViT-S/16
        & 55.63
        & -- \\

        \midrule
        \multicolumn{5}{l}{\textit{RGB Initialization$^{\dagger}$}} \\
        BEiT
        & ICLR'22
        & ViT-B/16
        & 53.10
        & -- \\

        MAE
        & CVPR'22
        & ViT-B/16
        & 67.68
        & -- \\

        MoCo-v3
        & ICCV'21
        & ViT-S/16
        & 76.59
        & -- \\

        DINOv2
        & TMLR'24
        & ViT-S/16
        & 91.94
        & 98.12 \\

        \midrule
        \multicolumn{5}{l}{\textit{Event Representation Learning$^{\dagger}$}} \\
        EventPillars
        & AAAI'25
        & ResNet-34
        & 85.30
        & -- \\

        EVA-L
        & ICLR'26
        & EVA-L
        & 86.30
        & -- \\

        OmniEvent
        & AAAI'26
        & ResNet-34
        & 90.20
        & -- \\

        \midrule
        \multicolumn{5}{l}{\textit{Event Pretraining$^{\dagger}$}} \\
        ECDP
        & ICCV'23
        & ViT-S/16
        & 87.66
        & -- \\

        EventBind
        & ECCV'24
        & ViT-B/16
        & 94.08
        & -- \\

        GEP
        & CVPR'26
        & ViT-B/16
        & 96.47
        & 99.56 \\

        \midrule
        Ours
        & --
        & ViT-S/16
        & 94.81
        & 98.83 \\

        Ours
        & --
        & ViT-B/16
        & \underline{97.62}
        & \underline{99.68} \\

        \textbf{Ours}
        & --
        & ViT-L/16
        & \textbf{98.25}
        & \textbf{99.89} \\

        \bottomrule
    \end{tabular}
    \endgroup
    \caption{Object recognition results on N-Caltech101. Top-1 ($\mathrm{acc1}$) and top-5 ($\mathrm{acc5}$) accuracies are reported.}
    \label{tab:object_recognition_results}
\end{table}

\begin{table*}[t]
    \centering
    \begingroup
    \small
    \renewcommand{\arraystretch}{1.0}
    \setlength{\tabcolsep}{2pt}

    \begin{tabular*}{\textwidth}{
        @{\extracolsep{\fill}}
        llcccccccccccc
        @{}
    }
        \toprule
        \multirow{2}{*}{Method}
        & \multirow{2}{*}{Backbone}
        & \multicolumn{6}{c}{MVSEC-Depth}
        & \multicolumn{6}{c}{DSEC-Depth} \\
        \cmidrule(lr){3-8}
        \cmidrule(lr){9-14}
        & & LP & 1\% & 5\% & 10\% & 20\% & Full
        & LP & 1\% & 5\% & 10\% & 20\% & Full \\
        \midrule

        DepthAnyEvent-R
        & ViT-S/16
        & 7.473
        & 7.542
        & 7.261
        & 6.794
        & 6.637
        & 6.465
        & 10.584
        & 10.347
        & 9.898
        & 9.534
        & 9.065
        & 8.880 \\

        ScaleEvent
        & ViT-S/16
        & 6.756
        & 6.930
        & 6.712
        & 6.477
        & 6.352
        & 6.145
        & 4.861
        & 4.983
        & 4.751
        & 4.728
        & 4.694
        & 4.564 \\

        \midrule

        Ours
        & ViT-S/16
        & \textbf{4.020}
        & \textbf{4.223}
        & \textbf{3.945}
        & \textbf{3.727}
        & \textbf{3.651}
        & \textbf{3.596}
        & \textbf{4.860}
        & \textbf{4.923}
        & \textbf{4.726}
        & \textbf{4.684}
        & \textbf{4.632}
        & \textbf{4.513} \\

        \bottomrule
    \end{tabular*}
    \endgroup

    \caption{Monocular depth estimation results on MVSEC-Depth and DSEC-Depth under linear probing (LP), few-shot fine-tuning, and full supervision (Full). Performance is measured by $\mathrm{RMSE}$, where lower values are better.}

    \label{tab:depth_linear_fewshot}
\end{table*}

\noindent\textbf{Feature Encoders.}
We evaluate DINOv3~\cite{simeoni2025dinov3} ViT-S/B/L encoders with $16 \times 16$ patches. At each scale, the image teacher and event encoder share the same architecture and pretrained initialization; the teacher is frozen, while the event encoder is optimized.

\noindent\textbf{Benchmark Setup.}
Following ScaleEvent~\cite{chen2026scaling}, we adopt the same task-specific decoders, training pipelines, and evaluation protocols for full supervision, linear probing (LP), and few-shot fine-tuning. LP freezes the encoder and optimizes only the decoder, while few-shot evaluation uses $1\%$, $5\%$, $10\%$, or $20\%$ of the annotations selected by fixed-interval sampling.

\noindent\textbf{Implementation Details.}
We optimize the event encoder with AdamW using a learning rate of $5 \times 10^{-6}$, weight decay of $10^{-4}$, gradient clipping at $0.1$, and exponential decay of $0.9$. We construct $128$ hyperedges, each containing an anchor and its top-$32$ related tokens. Training runs for $15$ epochs on $4$ NVIDIA RTX 3090 GPUs with $100{,}000$ pairs per epoch. Additional details are provided in the \textbf{\textit{Appendix}}.

\subsection{Depth Estimation}

\noindent\textbf{Settings.}
We employ DAv2~\cite{yang2024depth} as the downstream depth decoder and initialize it with the released pre-trained weights. Following DepthAnyEvent~\cite{bartolomei2025depth}, we report absolute relative error ($\mathrm{AbsRel}$), root mean squared error ($\mathrm{RMSE}$), logarithmic RMSE ($\mathrm{RMSE}_{\log}$), and threshold accuracies $\delta_1$, $\delta_2$, and $\delta_3$, corresponding to $1.25$, $1.25^2$, and $1.25^3$, respectively. Lower error and higher accuracy indicate better performance.

\noindent\textbf{Results.}
Table~\ref{tab:event_depth_results} reports the monocular depth estimation results. The ViT-L/16 variant achieves the strongest overall performance on both benchmarks. On MVSEC-Depth, it improves over ScaleEvent by $0.182$, $0.113$, and $0.049$ in $\delta_1$, $\delta_2$, and $\delta_3$, respectively, while reducing $\mathrm{AbsRel}$, $\mathrm{RMSE}$, and $\mathrm{RMSE}_{\log}$ by $0.116$, $2.171$, and $0.141$. These gains across both accuracy and error metrics indicate that the learned representations encode effective geometric information. On DSEC-Depth, the same model increases $\delta_1$ from $0.896$ to $0.904$ and reduces $\mathrm{AbsRel}$ from $0.101$ to $0.096$ and $\mathrm{RMSE}$ from $3.694$ to $3.624$. 

\subsection{Object Recognition}

\noindent\textbf{Settings.}
Following GEP~\cite{cao2026generative}, we employ a multilayer perceptron (MLP) as the downstream classification head and report top-$1$ accuracy ($\mathrm{acc1}$) and top-$5$ accuracy ($\mathrm{acc5}$) as the evaluation metrics.

\noindent\textbf{Results.}
Table~\ref{tab:object_recognition_results} reports recognition results on N-Caltech101. With ViT-S/16, Hyper-RED achieves $94.81\%$ top-$1$ accuracy, exceeding the matched-scale DINOv2~\cite{oquab2023dinov2} and ECDP~\cite{yang2023event} baselines by $2.87$ and $7.15$ percentage points, respectively. With ViT-B/16, it reaches $97.62\%$ top-$1$ and $99.68\%$ top-$5$ accuracy, surpassing GEP~\cite{cao2026generative} by $1.15$ and $0.12$ percentage points under the same backbone scale. Increasing the encoder to ViT-L/16 raises the two scores to $98.25\%$ and $99.89\%$, establishing state-of-the-art performance.

\subsection{Semantic Segmentation}

\begin{table}[t]
    \centering
    \begingroup
    \small
    \renewcommand{\arraystretch}{1.05}
    \setlength{\tabcolsep}{1.1pt}

    \begin{tabular*}{\columnwidth}{
        @{\extracolsep{\fill}}
        lllcccc
        @{}
    }
        \toprule
        \multirow{2}{*}{Method}
        & \multirow{2}{*}{Venue}
        & \multirow{2}{*}{Backbone}
        & \multicolumn{2}{c}{DDD17}
        & \multicolumn{2}{c}{DSEC} \\
        \cmidrule(lr){4-5}
        \cmidrule(lr){6-7}
        & & &
        Acc$\uparrow$ &
        mIoU$\uparrow$ &
        Acc$\uparrow$ &
        mIoU$\uparrow$ \\
        \midrule

        \multicolumn{7}{l}{
            \textit{RGB Initialization$^{\dagger}$}
        } \\

        Ev-SegNet
        & CVPRW'19
        & Xception
        & 89.76
        & 54.81
        & 88.61
        & 51.76 \\

        E2VID
        & TPAMI'19
        & ResNet-18
        & 85.84
        & 48.47
        & 80.06
        & 44.08 \\


        PVT-FPN
        & ICCV'21
        & ResNet-34
        & \underline{94.28}
        & 53.89
        & --
        & -- \\


        MaskCLIP
        & ECCV'22
        & ViT-B/16
        & 90.50
        & 61.27
        & 89.81
        & 55.01 \\

        ESS
        & ECCV'22
        & E2VID
        & 88.43
        & 53.09
        & 84.17
        & 45.38 \\

        ESS-Sup
        & ECCV'22
        & E2VID
        & 91.08
        & 61.37
        & 89.37
        & 53.29 \\

        HMNet
        & CVPR'23
        & HMNet-L1
        & --
        & --
        & 89.80
        & 55.00 \\

        EvSegformer
        & TIP'23
        & MiT-B1
        & \textbf{94.72}
        & 54.41
        & --
        & -- \\

        FC-CLIP
        & NeurIPS'23
        & CNeXt-L
        & 90.68
        & 62.01
        & 89.97
        & 55.67 \\

        DINOv2
        & TMLR'24
        & ViT-S/16
        & --
        & 53.85
        & --
        & 52.17 \\

        HALSIE
        & WACV'24
        & SNN-ANN
        & 92.50
        & 60.66
        & 89.01
        & 52.43 \\

        ESEG
        & AAAI'25
        & MiT-B1
        & 90.68
        & 59.97
        & 91.47
        & 57.55 \\

        KWYAF
        & AAAI'25
        & MiT-B0
        & 91.32
        & 62.41
        & 90.87
        & 57.75 \\

        \midrule
        \multicolumn{7}{l}{
            \textit{Event Pretraining$^{\dagger}$}
        } \\

        ECDP
        & ICCV'23
        & ResNet-50
        & --
        & 59.15
        & --
        & 59.16 \\

        ECDDP
        & ECCV'24
        & ViT-S/16
        & --
        & 55.73
        & --
        & 56.38 \\

        ECDDP
        & ECCV'24
        & Swin-T/7
        & --
        & 62.56
        & --
        & 61.25 \\

        OpenESS
        & CVPR'24
        & ResNet-50
        & --
        & 57.01
        & --
        & 55.01 \\

        OpenESS
        & CVPR'24
        & E2VID
        & 91.05
        & 63.00
        & 90.21
        & 57.21 \\

        STP
        & ICCV'25
        & ResNet-50
        & --
        & 62.13
        & --
        & 61.29 \\

        STP
        & ICCV'25
        & Swin-T/7
        & --
        & 63.29
        & --
        & 62.05 \\

        GEP
        & CVPR'26
        & ViT-B/16
        & --
        & 61.90
        & --
        & 67.37 \\

        ScaleEvent
        & CVPR'26
        & ViT-L/16
        & 92.62
        & 65.08
        & 93.10
        & \textbf{69.65} \\

        \midrule

        Ours
        & --
        & ViT-S/16
        & 91.58
        & 62.55
        & 91.83
        & 61.43 \\

        Ours
        & --
        & ViT-B/16
        & 92.54
        & \underline{65.58}
        & \underline{93.14}
        & 65.35 \\

        \textbf{Ours}
        & --
        & ViT-L/16
        & 93.08
        & \textbf{67.16}
        & \textbf{93.86}
        & \underline{69.04} \\

        \bottomrule
    \end{tabular*}
    \endgroup

    \caption{Semantic segmentation results on DDD17-Seg and DSEC-Semantic, with all metrics reported in percentage (\%).}
    \label{tab:semantic_segmentation_results}
\end{table}

\begin{table*}[t]
    \centering
    \begingroup
    \small
    \renewcommand{\arraystretch}{1.0}
    \setlength{\tabcolsep}{2pt}

    \begin{tabular*}{\textwidth}{
        @{\extracolsep{\fill}}
        llcccccccccccc
        @{}
    }
        \toprule
        \multirow{2}{*}{Method}
        & \multirow{2}{*}{Backbone}
        & \multicolumn{6}{c}{DDD17-Seg}
        & \multicolumn{6}{c}{DSEC-Semantic} \\
        \cmidrule(lr){3-8}
        \cmidrule(lr){9-14}
        & & LP & 1\% & 5\% & 10\% & 20\% & Full
        & LP & 1\% & 5\% & 10\% & 20\% & Full \\
        \midrule

        MaskCLIP
        & ViT-B/16
        & 31.91
        & 53.91
        & 56.27
        & 59.32
        & 59.97
        & 61.27
        & 33.08
        & 33.89
        & 37.03
        & 38.83
        & 42.40
        & 55.01 \\

        FC-CLIP
        & ConvNeXt-L
        & 54.07
        & 56.38
        & 58.50
        & 60.05
        & 60.85
        & 62.01
        & 43.00
        & 39.12
        & 43.71
        & 44.09
        & 47.77
        & 55.67 \\

        OpenESS
        & E2VID
        & 55.61
        & 57.58
        & 59.07
        & 61.03
        & 61.78
        & 63.00
        & 44.26
        & 41.41
        & 44.97
        & 46.25
        & 48.28
        & 57.21 \\

        ScaleEvent
        & ViT-B/16
        & 57.87
        & 57.23
        & 59.54
        & 61.45
        & 62.06
        & 62.81
        & 58.42
        & 54.37
        & \textbf{62.82}
        & \textbf{63.88}
        & 64.15
        & 64.93 \\

        \midrule

        Ours
        & ViT-B/16
        & \textbf{62.94}
        & \textbf{60.86}
        & \textbf{63.51}
        & \textbf{64.04}
        & \textbf{64.68}
        & \textbf{65.58}
        & \textbf{61.94}
        & \textbf{56.16}
        & 62.17
        & 63.65
        & \textbf{64.95}
        & \textbf{65.35} \\

        \bottomrule
    \end{tabular*}
    \endgroup

\caption{Semantic segmentation results on DDD17-Seg and DSEC-Semantic under linear probing (LP), few-shot fine-tuning, and full supervision (Full). All $\mathrm{mIoU}$ scores are reported in percentage (\%).}
    \label{tab:linear_fewshot_segmentation}
\end{table*}

\begin{table*}[t]
    \centering
    \small
    \setlength{\tabcolsep}{5.2pt}
    \renewcommand{\arraystretch}{1.08}
    \begin{tabular}{@{}ccccccccccccc@{}}
        \toprule
        \multirow{2}{*}{Exp.}
        & \multirow{2}{*}{Distill}
        & \multirow{2}{*}{PRD}
        & \multirow{2}{*}{HOI}
        & \multirow{2}{*}{HOC}
        & \multicolumn{2}{c}{MVSEC-Depth}
        & \multicolumn{2}{c}{DSEC-Depth}
        & \multicolumn{2}{c}{DDD17-Seg}
        & \multicolumn{2}{c}{DSEC-Semantic} \\
        \cmidrule(lr){6-7}
        \cmidrule(lr){8-9}
        \cmidrule(lr){10-11}
        \cmidrule(lr){12-13}
        & & & & &
        $\delta_1 \uparrow$ &
        RMSE $\downarrow$ &
        $\delta_1 \uparrow$ &
        RMSE $\downarrow$ &
        Acc. $\uparrow$ &
        mIoU $\uparrow$ &
        Acc. $\uparrow$ &
        mIoU $\uparrow$ \\
        \midrule

        (a)
        &
        &
        &
        &
        &
        0.593
        & 6.635
        & 0.846
        & 4.424
        & 91.39
        & 59.60
        & 91.94
        & 64.31 \\

        (b)
        & \checkmark
        &
        &
        &
        & 0.609
        & 6.114
        & 0.875
        & 4.063
        & 92.16
        & 62.41
        & 92.74
        & 66.17 \\

        (c)
        & \checkmark
        & \checkmark
        &
        &
        & 0.739
        & 4.238
        & 0.893
        & 3.713
        & 92.48
        & 65.82
        & 93.66
        & 68.73 \\

        (d)
        & \checkmark
        & \checkmark
        & \checkmark
        &
        & 0.792
        & 3.470
        & 0.896
        & 3.727
        & 92.72
        & 66.24
        & 93.84
        & 68.79 \\

        (e)
        & \checkmark
        & \checkmark
        &
        & \checkmark
        & 0.799
        & 3.448
        & 0.893
        & 3.734
        & 92.61
        & 67.01
        & 93.81
        & 68.68 \\

        (f)
        & \checkmark
        & \checkmark
        & \checkmark
        & \checkmark
        & \textbf{0.807}
        & \textbf{3.383}
        & \textbf{0.904}
        & \textbf{3.624}
        & \textbf{93.08}
        & \textbf{67.16}
        & \textbf{93.86}
        & \textbf{69.04} \\
        \bottomrule
    \end{tabular}
    \caption{Component ablation using ViT-L/16 under full supervision. Distill, PRD, HOI, and HOC denote feature distillation, pairwise relational distillation, and the proposed high-order intra- and cross-modal losses, respectively. Exp. (a) uses image-domain initialization, and Exp. (f) represents the full model.}
    \label{tab:ablation}
\end{table*}

\noindent\textbf{Settings.} We employ EoMT~\cite{kerssies2025your} as the downstream semantic segmentation decoder and initialize it with the released pre-trained weights. For evaluation, following OpenESS~\cite{kong2024openess}, we report mean intersection over union ($\mathrm{mIoU}$) and accuracy ($\mathrm{Acc}$).

\begin{figure}[!b]    
    \centering\includegraphics[width = \linewidth]{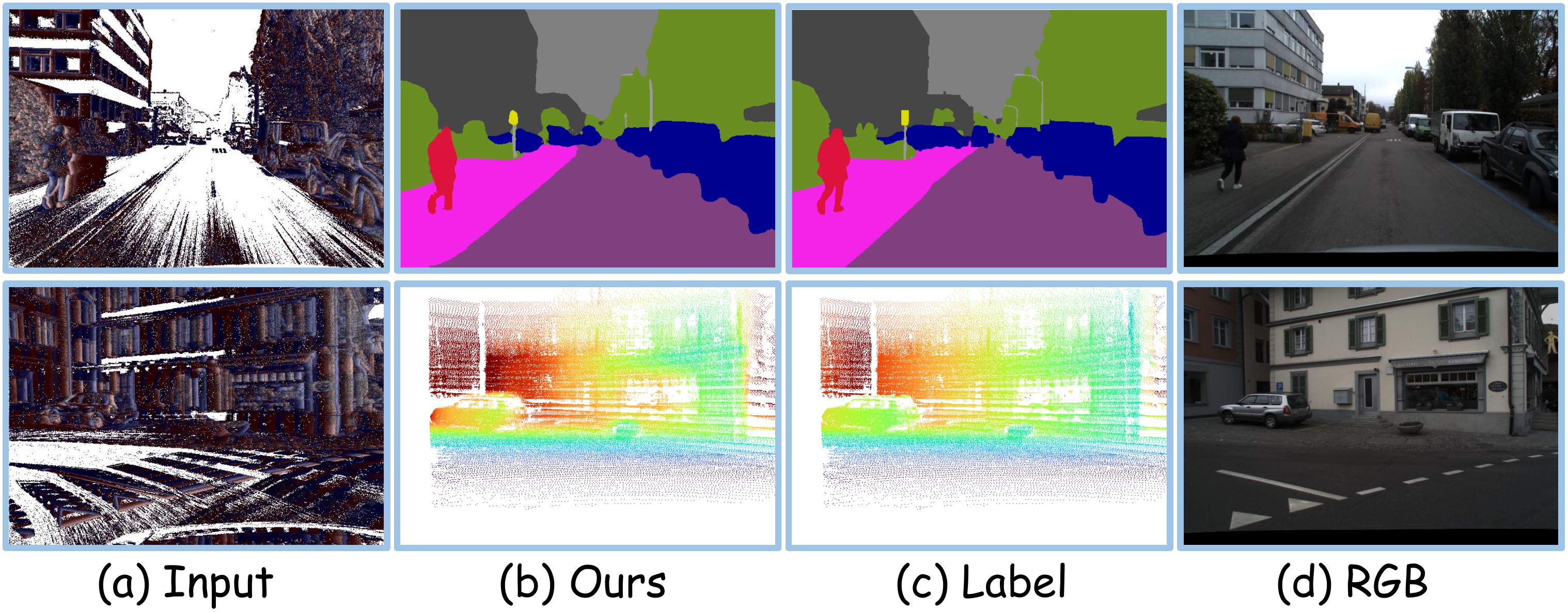}
    \caption{Qualitative results on semantic segmentation (top) and monocular depth estimation (bottom).}
    \label{fig:result}
\end{figure}

\noindent\textbf{Results.}
Table~\ref{tab:semantic_segmentation_results} presents the semantic segmentation results. On DDD17-Seg, our ViT-L/16 model achieves $67.16\%$ $\mathrm{mIoU}$, improving over ScaleEvent by $2.08$ percentage points. On DSEC-Semantic, it obtains the highest $\mathrm{Acc}$ of $93.86\%$ and the second-best $\mathrm{mIoU}$ of $69.04\%$, only $0.61$ percentage points below ScaleEvent. Under the same ViT-L/16 backbone, Hyper-RED improves $\mathrm{Acc}$ over ScaleEvent by $0.46$ and $0.76$ percentage points on DDD17-Seg and DSEC-Semantic, respectively. Figure~\ref{fig:result} further presents qualitative results on DSEC, showing accurate predictions for both semantic segmentation and monocular depth estimation. Additional qualitative results and visual comparisons with prior methods are provided in the \textbf{\textit{Appendix}}.

\subsection{Scalability Analysis}

We assess Hyper-RED along two practical dimensions: model-capacity scaling and annotation-efficient transfer.

\noindent\textbf{Model-Capacity Scaling.}
Using the same formulation and $16 \times 16$ patches, Hyper-RED improves monotonically across ViT-S/B/L on N-Caltech101 top-$1$ accuracy ($94.81\% \rightarrow 97.62\% \rightarrow 98.25\%$), DDD17-Seg $\mathrm{mIoU}$ ($62.55\% \rightarrow 65.58\% \rightarrow 67.16\%$), and MVSEC-Depth $\mathrm{RMSE}$ ($3.596 \rightarrow 3.449 \rightarrow 3.383$). These consistent gains across recognition, dense segmentation, and geometric prediction show that increased capacity translates into task-general improvements rather than benefiting a specific benchmark. Hyper-RED therefore scales to larger encoders without scale-specific modifications to its distillation formulation.

\noindent\textbf{Transferability and Annotation Efficiency.}
Tables~\ref{tab:depth_linear_fewshot} and~\ref{tab:linear_fewshot_segmentation} show that Hyper-RED outperforms ScaleEvent across all protocols on MVSEC-Depth and DDD17-Seg, reducing $\mathrm{RMSE}$ by $2.549$--$2.767$ and improving $\mathrm{mIoU}$ by $2.59$--$5.07$ percentage points, respectively. It also achieves the lowest DSEC-Depth $\mathrm{RMSE}$ throughout and ranks first under LP, 1\%, 5\%, 10\%, 20\%, and full-supervision settings, while remaining competitive otherwise. The strong LP and low-shot results indicate that the pretrained encoder already captures transferable semantic and geometric structures before extensive task-specific adaptation. Overall, Hyper-RED consistently benefits from larger encoders while remaining effective across diverse tasks and annotation budgets.

\subsection{Ablation Studies}

Table~\ref{tab:ablation} evaluates each component. Feature distillation improves over image-domain initialization, while PRD further establishes a stronger pairwise-relational baseline. Building on Exp.~(c), $\mathcal{L}_{\mathrm{HOI}}$ and $\mathcal{L}_{\mathrm{HOC}}$ yield task-dependent gains, reflecting their complementary roles: the former preserves event-token organization, whereas the latter anchors event groups to image semantics. Combining both in Exp.~(f) achieves the best results across all metrics. Relative to Exp.~(c), it improves MVSEC-Depth $\delta_1$ by $0.068$ and reduces RMSE by $0.855$, while increasing mIoU by $1.34$ and $0.31$ points on DDD17-Seg and DSEC-Semantic, respectively. These results validate the effectiveness and complementarity of the proposed high-order objectives. Additional ablation results and further analyses are provided in the \textbf{\textit{Appendix}}.

\section{Conclusion}

In this work, we presented Hyper-RED, a scalable image-to-event pre-training framework that addresses semantic collapse caused by rigid point-wise alignment across the image--event modality gap. Instead of forcing isolated token correspondences, Hyper-RED organizes semantically correlated tokens into teacher-guided hypergraphs and transfers their region-level organization through complementary intra- and cross-modal high-order relational distillation. By aligning both hyperedge memberships and prototypes, it preserves relational consistency and group-level semantics while retaining event-specific characteristics. Built on real and simulated pre-training data, Hyper-RED scales consistently from ViT-S to ViT-L without scale-specific modifications. Experiments on three downstream tasks across five event-based datasets demonstrate state-of-the-art overall performance and strong transferability under linear probing, few-shot fine-tuning, and full supervision. These results establish semantic hypergraphs as structured knowledge carriers for scalable and annotation-efficient event representation learning.

\bibliography{aaai2027}

\end{document}